\documentclass[10pt,twocolumn,letterpaper]{article}

\usepackage{cvpr}
\usepackage{times}
\usepackage{epsfig}
\usepackage{graphicx}
\usepackage{amsmath}
\usepackage{amssymb}

\usepackage{multirow}
\usepackage[T1]{fontenc}
\usepackage{colortbl}
\usepackage{multirow}
\usepackage{anyfontsize}
\usepackage{float}
\usepackage{xfrac}
\usepackage{caption}
\usepackage[ruled,vlined]{algorithm2e}
\usepackage{pstricks}
\usepackage{psfrag}

\newcommand{\bZero}{{\mathbf{0}}}

\newcommand{\bb}{{\mathbf{b}}}
\newcommand{\bc}{{\mathbf{c}}}

\newcommand{\bx}{{\mathbf{x}}}
\newcommand{\by}{{\mathbf{y}}}
\newcommand{\bz}{{\mathbf{z}}}

\newcommand{\bI}{{\mathbf{I}}}

\newcommand{\bbR}{{\mathbb{R}}}

\newcommand{\bbH}{{\mathbb{H}}}

\newcommand{\blambda}{\mbox{\boldmath $\lambda$}}

\newcommand{\bepsilon}{\mbox{\boldmath $\epsilon$}}

\def\argmin{\mathop{argmin}\limits}
\def\argmax{\mathop{argmax}\limits}
\def\argmin2{\mathop{argmin}\nolimits}
\def\argmax2{\mathop{\arg\max}\nolimits}

\newcommand{\comment}[1]{}

\usepackage[pagebackref=true,breaklinks=true,letterpaper=true,colorlinks,bookmarks=false]{hyperref}
\cvprfinalcopy 

\begin{document}

\title{Implicit Sparse Code Hashing}

\author{Tsung-Yu Lin\hspace{1.00cm} Tsung-Wei Ke \hspace{1.00cm} Tyng-Luh Liu\\
Institute of Information Science\\
Academia Sinica, Taiwan
}

\maketitle
\thispagestyle{empty}

\begin{abstract}
   We address the problem of converting large-scale high-dimensional image data into binary codes so that approximate nearest-neighbor search over them can be efficiently performed. Different from most of the existing unsupervised approaches for yielding binary codes, our method is based on a dimensionality-reduction criterion that its resulting mapping is designed to preserve the image relationships entailed by the inner products of sparse codes, rather than those implied by the Euclidean distances in the ambient space. While the proposed formulation does not require computing any sparse codes, the underlying computation model still inevitably involves solving an unmanageable eigenproblem when extremely high-dimensional descriptors are used. To overcome the difficulty, we consider the column-sampling technique and presume a special form of rotation matrix to facilitate subproblem decomposition. We test our method on several challenging image datasets and demonstrate its effectiveness by comparing with state-of-the-art binary coding techniques.
\end{abstract}


%
\section{Introduction}
\label{sec:intro}
%

How to efficiently perform similarity search over a large-scale image database is an important research topic with rich applications in computer vision. Its recent advances have notably led to two promising developments, namely, representing each image via a {\em high-dimensional descriptor} to improve the retrieval accuracy, and encoding it with a {\em binary code} to speed up the task of finding relevant ones. While such performance gains naturally call for efforts to connect the two advancements for achieving both accuracy and efficiency, there are major obstacles concerning the computation cost to be overcome in formulating a unified approach. Specifically, the large scale of an underlying image database and the high dimensions of the adopted features unavoidably induce enormous data matrices. These, in turn, could cause various numerical-challenging issues in computing the binary codes, which can either exceed the capacity of available computing resources or require time-consuming training and testing processes. Our goal is to establish an effective framework for yielding binary codes so that similarity search over large-scale high-dimensional image data can be conveniently carried out.

We consider {\em unsupervised} binary coding to account for that in practical applications involving large-scale image data, the label information is often not completely available. However, the lack of {\em label ground truth} imposes extra difficulty on evaluating the retrieval accuracy. To resolve this matter, previous techniques use a nominal threshold based on the average Euclidean distance to the $k$th-nearest neighbor of each image to form the so-called {\em Euclidean ground truth} \cite{RaginskyL09}. While this strategy is handy to provide a universal criterion for quantifying the retrieval results, it nevertheless brings in new concerns. First, when comparing, say, two binary coding schemes, it is questionable that the one yielding better retrieval performance on the Euclidean ground truth would guarantee the same advantage on the label ground truth (when available). Second, adopting the Euclidean ground truth implicitly implies that performing similarity search based on any resulting binary coding methods would not outperform using $2$-norm in the original input space. We instead evaluate the performance according to the label ground truth. Such a choice enables less biased comparisons and more likely better generalizations. Still, we emphasize that the label ground truth is not needed for the proposed formulation but solely for achieving more reliable evaluations on the experimental results.

Exploring sparse codes has been shown to be advantageous in many relevant studies. Gkioulekas and Zickler \cite{GkioulekasZ11} have introduced an unsupervised dimensionality reduction framework by preserving the pairwise inner products of sparse codes. Their approach can improve recovering meaningful image relationships for different classes of signals, including facial images, texture patches, and images of object categories. Timofte and Van Gool \cite{timofte2011sparse} show that by defining the affinity matrix based on sparse codes, the effectiveness of the resulting embedding can be enhanced. The method in \cite{CherianSMP14} exploits sparse codes to improve nearest-neighbor search, and focuses on learning the dictionary and sparse codes robustly. The derived sparse codes are then mapped to hash codes, which are used for indexing to a hash table for similarity search. Inspired by these relevant improvements, we establish our approach to binary coding by investigating the image structure implied by sparse codes. However, owing to the technique in \cite{GkioulekasZ11}, our method distinctly does not involve an explicit computation of sparse codes and is thus favorable for dealing with large-scale data.

%
\section{Related work}
\label{sec:related}
%
%

Choosing a proper feature representation to describe an image is crucial to the success of most computer vision techniques. This is especially true in dealing with large-scale problems. Most of the existing high-dimensional descriptors are typically designed to encode as many aspects of image statistics as possible. Take, for example, the Fisher Vector (FV) representation \cite{Perronnin2007fisher, Perronnin2010improving}. It is established by assuming that local features are generated from a Gaussian Mixture Model (GMM), which can be learned from maximum likelihood estimation. By differentiating the log-likelihood of local features with respect to one type of the GMM parameters, one can derive a specific form of FV. The often-used Bag-of-Words (BoW) model can be thought of as a simple case of FV. In \cite{Sanchez2011high}, S\'{a}nchez and Perronnin demonstrate that FV is effective for large-scale classification problems and report significant improvements over then state-of-the-art on ILSVRC 2010 \cite{Berg2010ilsvrc}. The application of FV has also been expanded to deal with object detection \cite{Cinbis2013segmentation} as well as action and event recognition \cite{Oneata2013action}. Another development closely related to FV is the introduction of Vector of Locally Aggregated Descriptors (VLAD), which is independently proposed by J\'{e}gou \etal \cite{Jegou2010aggregating}. The way VLAD models the distribution of local image features is similar to the Fisher kernel representation \cite{Perronnin2007fisher}. VLAD is thus a non-probabilistic simplification of FV. Arandjelovi\'{c} and Zisserman \cite{ArandjelovicZ13} have provided a thorough study about VLAD and proposed useful techniques to explore VLAD more effectively. The Locality constrained Linear Coding (LLC) \cite{Wang2010locality} is also a high-dimensional descriptor where its coding process can be pictured as a hierarchical architecture of carrying out feature extraction, sparse coding, pooling and feature vector concatenation according to the spatial layout of Spatial Pyramid Matching (SPM) \cite{Lazebnik2006beyond}.

For unsupervised binary coding, Locality Sensitive Hashing (LSH) first formalizes the concept of approximate nearest-neighbor search to find, in high probability and in sub-linear time, items within $(1+\epsilon)$ times the optimal similarity \cite{GionisIM99}. As LSH relies on random projections, it often requires a long code to achieve good precision. Weiss \etal \cite{WeissTF08} propose a hard criterion, which is related to {\em graph partitioning}, to establish Spectral Hashing (SH). However, SH assumes that the data are generated from a separable distribution, and seems not to be competitive as reported in \cite{GongL11}. Subsequently, the same group of authors have extended SH to MultiDimensional Spectral Hashing (MDSH) by considering a different criterion that matches Hamming affinity with the target affinity matrix. In \cite{RaginskyL09}, Raginsky and Lazebnik introduce a distribution-free hashing method (termed as SKLSH), which has an encoding scheme similar to SH, but instead uses randomly sampled directions. It is shown that,  as opposed to SH, SKLSH does not display the degenerate behavior with respect to the code length. ITerative Quantization (ITQ) by Gong and Lazebnik \cite{GongL11} is an effective framework for deriving similarity-preserving binary codes. It works by first using PCA to reduce the dimension of the data to the desired code size, and then carrying out an alternating optimization strategy to find the optimal rotation and binary codes  minimizing the quantization error. In \cite{HeS13}, He \etal have proposed K-Means Hashing (KMH) for binary codes. The idea is to generalize K-means clustering to iteratively minimize the quantization error and the affinity error in the alternating EM steps. In this respect, KMH can be linked to ITQ. SPherical Hashing (SPH) is introduced in \cite{HeoY12}. Rather than partitioning the data with respect to hyperplanes, SPH uses hyperspheres and measures the resulting binary codes with the spherical Hamming distance.

While the aforementioned binary coding methods are not designed to handle high-dimensional data, Bilinear Projection-based Binary Coding (BPBC) by Gong \etal \cite{GongKRL13} is the first attempt to explicitly tackle the challenging issue. BPBC can be considered a high-dimensional extension to ITQ. It is formulated to seek a proper bilinear projection to reduce the complexity of computing a full rotation and to achieve dimensionality reduction simultaneously. In \cite{Yu2014}, Yu \etal subsequently develop Circulant Binary Embedding (CBE). The algorithm works by imposing a circulant structure on the dimensionality-reduction projection matrix so that techniques based on fast Fourier transform can be applied to compute binary codes more efficiently. Xia \etal \cite{xia2015sparse} propose a Sparse Projection-based (SP) binary coding scheme for high-dimensional data. They consider a sparsity regularizer to achieve efficiency in both storage and encoding. However, their method is established upon first solving an eigenproblem and thus does not scale well to the data dimension. The length of an SP binary code can be smaller or greater than the original feature dimension. In the former case, the performance gradually improves to that of ITQ when increasing the code length; in the latter it gives similar performances as LSH \cite{xia2015sparse}.

It can be observed that the three high-dimensional binary coding methods, namely, BPBC, CBE and SP, are motivated by ITQ. As the retrieval performances of unsupervised binary codes are evaluated typically based on the Euclidean ground truth, it is insightful to note that their improvements over ITQ are generally not obvious. When the feature dimension is too large to run ITQ, only BPBC and CBE are applicable. The stagnation of achieving significant progress among the newly proposed techniques is mostly due to that the performance by ITQ (when using sufficient bits) is already rather close to using 2-norm to carry out similarity search in the original space. On the other hand, when testing these algorithms with label ground truth, their performance may not even surpass ITQ. Thus, for the sake of achieving more meaningful evaluations, we compare the various binary coding schemes solely based on label ground truth.

Our approach to binary coding is to preserve  the inner products of sparse codes, where the advantages of respecting such quantities are justified in, \eg, \cite{CherianSMP14,GkioulekasZ11,timofte2011sparse}. Since the sparse codes are not explicitly computed, we term our binary coding method as Implicit Sparse Code Hashing (ISCH). The first step of ISCH is to learn a dictionary $D$ from the image database and (conceptually) represent each image by its sparse code. It then carries out dimensionality reduction to preserve the image relationships entailed by the inner products between sparse codes. Leveraging with the techniques in \cite{GkioulekasZ11,Seeger08}, all the sparse codes are not needed to be computed at all. Otherwise, the computation time caused by the process of sparse coding would make ISCH unfeasible for large-scale data. When the feature dimension is extremely large, the main challenge in completing the reduction step is to solve the eigenproblem for $DD^T$. To tackle the high complexity, we consider the {\em column sampling} low-rank approximation methods proposed in \cite{Kumar2009sampling,Mackey2011divide}. We then approximately solve the large number of eigenvalues and the corresponding eigenvectors. Finally, the resulting mapping to generate binary codes is given by a closed-form formula up to an arbitrary rotation, which in turn can be decided by minimizing the quantization error and by assuming a special structure of the rotation matrix.

%
\section{Implicit sparse code hashing}
\label{sec:ISCH}
%

Given a collection of $n$ images, denoted as $\Omega = \{\bx_i\in \bbR^d\}_{i=1}^n$, our task is to convert each $\bx_i$ into an $m$-bit ($m < d$) binary code $\bb_i \in \{0,1\}^m$ so that similarity search can be efficiently performed. We express the mapping by
\begin{equation}
\bx \in \bbR^d \longmapsto \bb \in \bbH^m
\label{eqn:Bmap}
\end{equation}
\noindent where $\bbH^m$ is the $m$-dimensional Hamming space. In our experiments, the value of $d$, the dimension of the image feature vector, can be up to 105,000.

As the motivation behind our binary code conversion is to consider the image neighborhood relationships implied by the sparse codes, we first need to learn a dictionary of $k$ atoms. We require $k > d$ so that $D$ can be overcomplete. To construct D, we adjust the image data to be zero-centered and apply an efficient implementation of k-means clustering from the {\tt Yael} library \cite{yael}. (More details about $D$ are given in Section~\ref{sec:exp}.) Each atom in $D$ is then normalized to be unit length. The sparse code $\bc_i$ of each $\bx_i \in \Omega$ is the solution to the following lasso problem:
\begin{equation}
\min_{\bc} \; \frac{1}{2} \| \bx - D \bc\|^2_2 + \eta \|\bc\|_1
\label{eqn:lasso}
\end{equation}
\noindent where $\eta$ is a parameter to weigh the influence of the sparseness prior. We denote the relation between an image and its sparse code as
\begin{equation}
\bx \in \bbR^d \longmapsto \bc \in \bbR^k.
\label{eqn:Cmap}
\end{equation}

\subsection{Dimensionality reduction}
\label{ssec:DR}

To obtain $m$-bit binary codes for $\Omega$, we carry out dimensionality reduction from $\bbR^d$ to $\bbR^m$. The linear mapping $L$ realizing the reduction is specified by
\begin{equation}
\bx \in \bbR^d \longmapsto L \bx = \by\in \bbR^m.
\label{eqn:Lmap}
\end{equation}
\noindent In most of the existing unsupervised binary coding methods, the mapping $L$ is decided by preserving the $2$-norm image relationships in $\bbR^d$. That is, $L$ is obtained by minimizing
\begin{equation}
\sum_{i \neq j} (\|\bx_i - \bx_j\|_2 - \|\by_i - \by_j\|_2)^2.
\label{eqn:Euclidean}
\end{equation}

\noindent However, using $2$-norm (\ie, Euclidean distance) to measure the similarity between images would often yield improper results and consequently incorrect image relations. The criterion in (\ref{eqn:Euclidean}) in some sense limits the effectiveness of its resulting binary codes. Motivated by \cite{GkioulekasZ11}, the proposed ISCH instead decides $L$ by respecting the inner products of sparse codes and minimizing the expected squared difference in inner product, \ie,
\begin{equation}
L^* =  \mathop{\arg}\min\limits_L \, \mathbf{E}_{\bc_i, \bc_j, \bx_i, \bx_j} \left[(\bc_i^T \bc_j - \by_i^T \by_j)^2\right].
\label{eqn:innerproduct}
\end{equation}

Gkioulekas and Zickler \cite{GkioulekasZ11} prove that without the need to compute the sparse codes by solving (\ref{eqn:lasso}), the optimization problem (\ref{eqn:innerproduct}) can be reduced to minimizing an objective function whose stationary points could be analytically obtained if the following {\em sparse linear model} \cite{Seeger08} is assumed:
\begin{equation}
\bx = D\,\bc + \bepsilon
\label{eqn:linear}
\end{equation}
\noindent
where  $\bepsilon \sim N(\bZero, \sigma^2 \bI)$ is the Gaussian noise of variance $\sigma^2$ and sparse code $\bc =(c_1, \dots, c_k)^T$ obeys the independent Laplace distribution priors,
\begin{equation}
P(\bc)  = \prod_{j=1}^k P(c_j), \; P(c_j) = \frac{1}{2\tau}\, \mathrm{exp}\left\{-\frac{|c_j|}{\tau}\right\}\,,
\label{eqn:Laplace}
\end{equation}
\noindent where $\tau$  is the positive scale parameter. Note that the weighting parameter in (\ref{eqn:lasso}) relates to (\ref{eqn:linear}) and (\ref{eqn:Laplace}) by $\eta = \sigma^2/\tau$. It follows from \cite{GkioulekasZ11} that the optimal linear projection $L^*$ is determined (up to an $m \times m$ rotation matrix $R$) by the following closed-form,
\begin{equation}
L^* = R^T \, \mathbf{diag}(f(\blambda_m))\, V_m^T \equiv R^T W
\label{eqn:L}
\end{equation}
\noindent where $\blambda_m = (\lambda_1,\dots,\lambda_m)^T$ includes the $m$ largest eigenvalues of the matrix $DD^T$ and $V_m$ consists of the $m$ corresponding eigenvectors as columns. The expression $\mathbf{diag}(f(\blambda_m))$ in (\ref{eqn:L}) represents an $m\times m$ diagonal matrix whose $i$th diagonal entry is given by
\begin{equation}
f(\lambda_i) = \sqrt{\frac{4 \tau^4 \lambda_i}{\sigma^4+4\tau^2\sigma^2\lambda_i+4\tau^4\lambda_i^2}}\,.
\label{eqn:f}
\end{equation}
\noindent How to compute $W=\mathbf{diag}(f(\blambda_m))\, V_m^T$ and $R$ in (\ref{eqn:L}) will be discussed in the next two sections. Suffice it to say now that they can only be approximately decided for the high-dimensional case. Once we have obtained an {\em optimal} $L^*$,  to perform an on-line similarity search with respect to a novel image $\bx$ can be done as follows. Let $L^*_i$ be the $i$th row of $L^*$ and $\bb = (b_1, \dots, b_m)^T$ be the binary code of $\bx$. Then we have, for $1 \leq i \leq m$,
\begin{equation}
b_i = \left\{
\begin{array}{ll}
1, &\quad \mbox{if $L^*_i \bx \geq 0$,} \\
0, &\quad \mbox{otherwise.}
\end{array}
\right.
\label{eqn:Bsign}
\end{equation}
According to (\ref{eqn:innerproduct}), those images in $\Omega$ that have similar sparse codes to that of $\bx$ can now be retrieved through XOR bit-wise operations over binary codes.

\subsection{Approximate spectral decomposition}
\label{ssec:SVD}

To approximately solve the high-dimensional eigenprobelm for $DD^T$, we use {\em sampling without replacement} to select $m$ columns from $D$ to form a $d \times m$ rectangular matrix, say, $C$ and express its compact singular value decomposition by
\begin{equation}
C = U_C \, \Sigma_C \, V_C^T
\label{eqn:SVD}
\end{equation}
where $U_C \in \bbR^{d\times m}$, $\Sigma_C \in \bbR^{m\times m}$ and $V_C \in \bbR^{m\times m}$. The rank of $C$ is assumed to be $m$. Otherwise, we adjust the column sampling to ensure this condition. (An efficient check is described below.) Let the Moore-Penrose pseudoinverse of $C$ be $C^{+} = V_C \Sigma_C^{-1} U_C^T$. We then carry out {\em column projection} to generate a ``matrix approximation'' of $D$ by
\begin{equation}
D_C^{\mathrm{proj}} = C C^+ D = U_C U_C^T D
\end{equation}
\noindent where $D_C^{\mathrm{proj}}$ can be thought of as an $m$-rank approximation to $D$ onto the column space of $C$.

Since $m$, the length of binary code, can be large, solving SVD for $D_C^{\mathrm{proj}}$ is still unmanageable in most cases. We further decompose it into $Q$ subprobelms. Specifically, we randomly select $\ell$ columns from $C$ and divide it into $Q$ sub-matrices of size $d \times \ell$, denoted as $\{C_i\}_{i=1}^Q$. (Assume that $\ell = m/Q$.) The decomposition is to enable a feasible SVD computation for each sub-matrix, which can now be expressed by $C_i=U_{C_i} \Sigma_{C_i} V_{C_i}^T$, for $i=1, \dots, Q$. If any $C_i$ does not have $\ell$ nonzero singular values, we redo the sampling without replacement for its corresponding subset of columns in $C$. We then approximate the SVD for $D_C^{\mathrm{proj}}$ by simultaneously solving $Q$ problems of SVD for
\begin{equation}
D_{C_i}^{\mathrm{proj}} = C_i C_i^+ D = U_{C_i} U_{C_i}^T D,\quad \mbox{i=1,\dots, Q}.
\label{eqn:Tproblems}
\end{equation}
\noindent Although $D_{C_i}^{\mathrm{proj}}$ is still a $d \times k$ matrix, its rank is only $\ell$. To accomplish SVD for $D_{C_i}^{\mathrm{proj}}$, we let $A_i = U_{C_i}^T D \in \bbR^{\ell \times k}$ and decompose it by $A_i = U_{A_i} \Sigma_{A_i} V_{A_i}^T$. (The SVD of $A_i$ can be efficiently computed owing to $\ell \ll k$.) Since the columns in $U_{C_i}U_{A_i}$ are orthogonal to each other, it follows that we have indeed derived the SVD of $D_{C_i}^{\mathrm{proj}}$ by
\begin{equation}
D_{C_i}^{\mathrm{proj}} = U_{C_i}A_i = U_{C_i}U_{A_i} \Sigma_{A_i} V_{A_i}^T,\quad \mbox{i=1,\dots, Q}.
\end{equation}
\noindent We thus obtain $\ell$ pairs of singular value and left singular vector for each $D_{C_i}^{\mathrm{proj}}$. The validity of transforming into $Q$ smaller SVD problems depends on how well the following approximation holds:
\begin{equation}
U_C U_C^T D \approx \sum_{i=1}^Q U_{C_i} U_{C_i}^T D
\label{eqn:Tsum}
\end{equation}
\noindent which is certainly a crucial topic for further study. Once we have collected the $m$ pairs of singular value and left singular vector of $D$, we thus approximately solve the eigenproblem for $DD^T$ and also the linear projection $W$ in (\ref{eqn:L}).

\subsection{Rotation matrix}
\label{ssec:R}

We are left to determine the rotation matrix $R \in \bbR^{m\times m}$ in (\ref{eqn:L}) to complete the linear projection $L^*=R^T W$. Since our goal is to seek a scheme for producing good binary codes, it is convenient to link the criterion for an appropriate $R$ to facilitating the desired encoding. Analogous to ITQ, we consider deriving the rotation matrix $R$ by minimizing the quantization error. That is, the optimal binary codes for all $\bx_i \in \Omega$ and the optimal $R$ are jointly optimized as follows:
\begin{equation}
\left\{R^*, \{\bb_i^*\} \right\} =  \mathop{\arg}\min\limits_{R, \{\bb_i\}} \sum_{i=1}^n \|\bb_i - R^T W \bx_i\|_2^2
\label{eqn:rotation}
\end{equation}
where $\bb_i \in \bbH^m$ is the $m$-bit binary code of $\bx_i$ as in (\ref{eqn:Bmap}).

Since the value of $m$ is generally quite large for high-dimensional data, directly solving (\ref{eqn:rotation}) may not be feasible. In BPBC \cite{GongKRL13}, the idea to restrict $R$ to a bilinear rotation has been discussed for reducing the computation complexity. In our approach, we simplify the large-scale optimization problem (\ref{eqn:rotation}) by assuming $R$ to be a sparse block-diagonal matrix. We have
\begin{equation}
R=\left[ {\begin{array}{cccc} R_1& & & \\ &R_2& & \\ & & \ddots & \\ & & &R_Q\\  \end{array}}\right]\in \bbR^{m\times m}
\label{eqn:block}
\end{equation}
where nonzero elements in $R$ appear only at the diagonal blocks $R_j \in \bbR^{\ell \times \ell}$, and  $m = Q \times \ell$ as before. With (\ref{eqn:block}), the optimization problem (\ref{eqn:rotation}) is reduced to
\begin{equation}
\left\{R^*, \{\bb_i^*\} \right\} =  \mathop{\arg}\min\limits_{R, \{\bb_i\}}  \sum_{j=1}^Q \sum_{i=1}^n \|\bb_i(j) - R_j^T\, \bz_i(j)\|_2^2
\label{eqn:rotation1}
\end{equation}
\noindent where we have divided each $\bb_i$ and $W\bx_i$ into $Q$ segments and used the notations $\bb_i(j)$ and $\bz_i(j)$ to represent their $j$th segment. Again, we have decomposed a large-scale problem into $Q$ smaller ones, each of which can be independently solved using an iterative process of alternating optimization. Namely, we fix $\bb_i(j)$ or $R_j$ in turn and optimize the other until a preset stopping criterion is met.

%
\section{Experimental results}
\label{sec:exp}
%

We carry out extensive experiments to evaluate the effectiveness of ISCH with five benchmark datasets, including CIFAR-10~\cite{TorralbaFF08}, MNIST \cite{LeCun726791}, ILSVRC2010~\cite{Deng2009imagenet}, INRIA Holiday+Flickr1M~\cite{Jegou2008hamming} and WDRef+LFW~\cite{ChenCWW012, LFWTech}. The choice of these datasets is to demonstrate that our method works for image descriptors of both high and moderate dimensions. For the sake of achieving more conclusive evaluations, the retrieval results are measured according to the label ground truth. The methods to be compared with include LSH, SPH, ITQ, BPBC, CBE, SP and using $2$-norm in the ambient space. As we will detail later that our method significantly outperforms the others in all the experiments, comprising datasets of different scales and feature dimensions. In particular, our results show that ISCH could use less bits than all other compared techniques to achieve comparable performance. Such an advantage is useful for both reducing the coding storage of an underlying large-scale image database and the testing time for carrying out exhaustive Hamming distance computation over the whole dataset.

\subsection{Evaluation protocols}
\label{ssec:protocal}

CIFAR-10 is a subset of Tiny Images dataset~\cite{TorralbaFF08}. It consists of 60,000 color images of size $32 \times 32$ pixels. Each of the ten categories in CIFAR-10 contains 6,000 images. We randomly select 1,000 images as test queries, 100 images from each class, and use all the remaining images to make up the training set. To represent an image in CIFAR-10, we use the grayscale GIST descriptor to form a $512$-D feature vector. For CIFAR-10, we construct a dictionary of size 1,024 in our ISCH method.

MNIST includes 70,000 $28 \times 28$ hand-written digit images, each of which corresponds to one of the digits from 0 to 9. As the image size is $28 \times 28$, it can be conveniently represented by a $784$-D vector of raw pixel values. We randomly sample 2,000 images, equally distributed in each category, as test queries and perform retrieval on the remaining image set. We generate 1,568 cluster centers by k-means algorithm to form an overcomplete dictionary.

ILSVRC2010, a subset of ImageNet~\cite{Deng2009imagenet}, is a challenging image collection for fine-grained category classification. It consists of 1.2M images corresponding to 1,000 classes. We download from the ImageNet website the public SIFT features. To construct the VLAD representation, we form 200 clusters and assign each SIFT descriptor to one of the clusters to represent an image. The resulting dimension is $128 \times 200 = 25600$. As suggested in~\cite{ArandjelovicZ13} that intra-normalization, \ie, the process to perform $2$-norm normalization over the sum of residuals within each VLAD cluster independently, would yield better performance than power-normalization, we thus normalize VLAD feature vectors by intra-normalization followed by $2$-norm normalization. To further evaluate the retrieval performances, we also consider the LLC~\cite{Wang2010locality} descriptor and the Caffe CNN-fc7 feature \cite{Caffe}. The setting for LLC is similar to that in \cite{GongKRL13}. We cluster the SIFT feature vectors to construct a codebook of size 5,000. For each image, the densely extracted SIFT descriptors are coded by LLC and aggregated by using three-level spatial pyramid and max pooling. This would yield a $5000 \times 21 = 105000$ dimensional representation. The LLC feature vectors are further processed by zero-centering and $2$-norm normalization. Regarding the CNN-fc7 representation, it is $4096$-D and the only {\em supervised} feature (obtained from pre-training with ImageNet) used in our experiments.

The INRIA Holiday database is a collection of personal holiday photos, including 1,491 images over 500 different scenes. 500 images from the dataset are randomly chosen as queries, while the remaining 991 images are combined with Flickr1M to form the underlying database. For each query, there are about two images to be retrieved from the database. We download from the author's website the public SIFT features and the dictionary with 500 vocabularies to construct a $128 \times 500 = 64000$ dimensional VLAD for encoding each image. Analogously, the resulting VLAD features are intra-normalized and then $2$-norm normalized.

WDRef+LFW is a mixed image collection that comprises the face database WDRef~\cite{ChenCWW012} and also LFW~\cite{LFWTech} serving as distractors.  For this dataset, we compare our method with only BPBC and 2-norm similarity search to highlight that ISCH for solving large-scale optimization is not restricted to the descriptor structure. WDRef consists of 71,846 images from 2,995 subjects, most of which have more than 10 images. We randomly sample one image per subject from WDRef to form a query set of 2,995 images. The remaining images are combined with LFW, which has 13,233 images, to form the underlying database. We represent each image with the LE~\cite{CaoYTS10} descriptor, which has an implicit $2$-D structure of dimensions
$20736 =256 \,\mbox{(local discriptor size)}\, \times 81 \,\mbox{($9\times 9$ patches)}$.
Additionally, we expand LE with a $5900$-D LBP to form the other high-dimensional descriptor. To balance the effects of LE and LBP, they are normalized to unit length independently and then concatenated. The combined vector is again normalized to unit length. One notable property of the LE+LBP descriptor is that the implicit $2$-D structure is no longer present. Since BPBC relies on exploring the $2$-D structure of a high-dimensional descriptor, it is not clear how to apply BPBC to work with the LE+LBP descriptor.

Details about how to evaluate the retrieval performance are as follows. The retrieval results of binary codes are decided by Hamming distance. With CIFAR-10 and MNIST, we use 2-norm similarity search in the original space as the baseline and test all methods except BPBC, which is more appropriate for high-dimensional data. With ILSVRC2010 and INRIA Holiday+Flickr1M, we focus on evaluating ITQ, BPBC, CBE, SP and ISCH. When the descriptor dimension is notably high, carrying out ITQ and SP would become too time-consuming and the two will be excluded from the evaluation. For the experiment with ILSVRC2010, we randomly sample 1,000 images as queries and assess the precision at top k nearest neighbors. For Holiady+Flickr1M, we adopt the measure used in ~\cite{Jegou2008hamming} and report mean average precision (mAP), \ie, the area under recall precision curve over 500 predefined queries. Finally, for WDRef+LFW, we also measure the top k nearest-neighbor performance. In all our experiments, the same set of query images are used for comparing the retrieval results respectively by Hamming distance and $2$-norm nearest-neighbor search.

\subsection{Implementation details}
\label{ssec:dicttrain}

When the feature dimension $d$ is large, learning a dictionary $D$ from a large-scale dataset is nontrivial and time-consuming. To efficiently accomplish the task, rather than directly performing k-means clustering in the original feature space, we simply project the data onto a lower-dimensional space where the resulting features of lower dimensions would serve as proxies to hierarchically cluster original features. Specifically, to hierarchically construct $D$ in H levels is carried out as follows. Let $X \in \bbR^{d\times n}$ be the data matrix. We first generate $H$ dimensionality-reduction random projections, namely, $P_1, \dots, P_H$ where $P_h \in \bbR^{d \times d_{\ell}}$, $h=1,\dots, H$ and $d_\ell \ll d$. At level $h$, the hierarchical clustering is to divide each cluster at level $h-1$ into $k_h$ clusters. To this end, we compute the level-wise reduction by $Y_h = P_h^T X \in \bbR^{d_\ell \times n}$. Then, k-means clustering is performed over the reduction images of each cluster. That is, we use $Y_h$ as proxy to reduce the complexity of clustering data in $X$. Since the cluster sizes can be imbalanced, a cluster of size smaller than some pre-specified threshold would not be further split into the next level. To form the dictionary, all the cluster centers produced through the hierarchical process are included as dictionary atoms. In our implementation, we empirically set $H=2$ and $k_2= 2 \times k_1$. That is to say, we generate at most $k_1+k_1 \times k_2=k_1(1+2k_1)$ atoms for a dictionary. The general principle of selecting the value of $k_1$ is to make sure that the size of dictionary is larger than $d$. In our experiment, we respectively set the value of $k_1$ to $128$, $256$, $200$ and $135$ for ILSVRC2010 with $25600$-D VLAD,  ILSVRC2010 with $105000$-D LLC, Holiday+Flickr1M with $64000$-D VLAD, and WDRef+LFW with $20736$-D LE and $26636$-D LE+LBP.

The key parameter in our method is $\tau$, the scale parameter of the Laplace distribution in (\ref{eqn:Laplace}). From (\ref{eqn:lasso}) and (\ref{eqn:Laplace}), the parameters of the Laplace distribution and the sparseness weight are related by $\sigma^2/\tau = \eta$. In \cite{YangYH10}, $\eta$ is suggested to be 0.05. We follow this setting and thus only need to decide the value of $\tau$ to specify the formula in (\ref{eqn:f}). As $\tau$ is the scale parameter, its suitable value correlates with the dictionary size. In our experiments, $\tau$ is set to $0.001$ for CIFAR-10 and MNIST, and $0.12$ for all others, except when using the CNN-fc7 feature, we have $\tau = 0.01$.

\begin{figure*}[tH]
\centering
\begin{tabular}{@{}c@{}c@{}c@{}}
\includegraphics[width=0.32\textwidth]{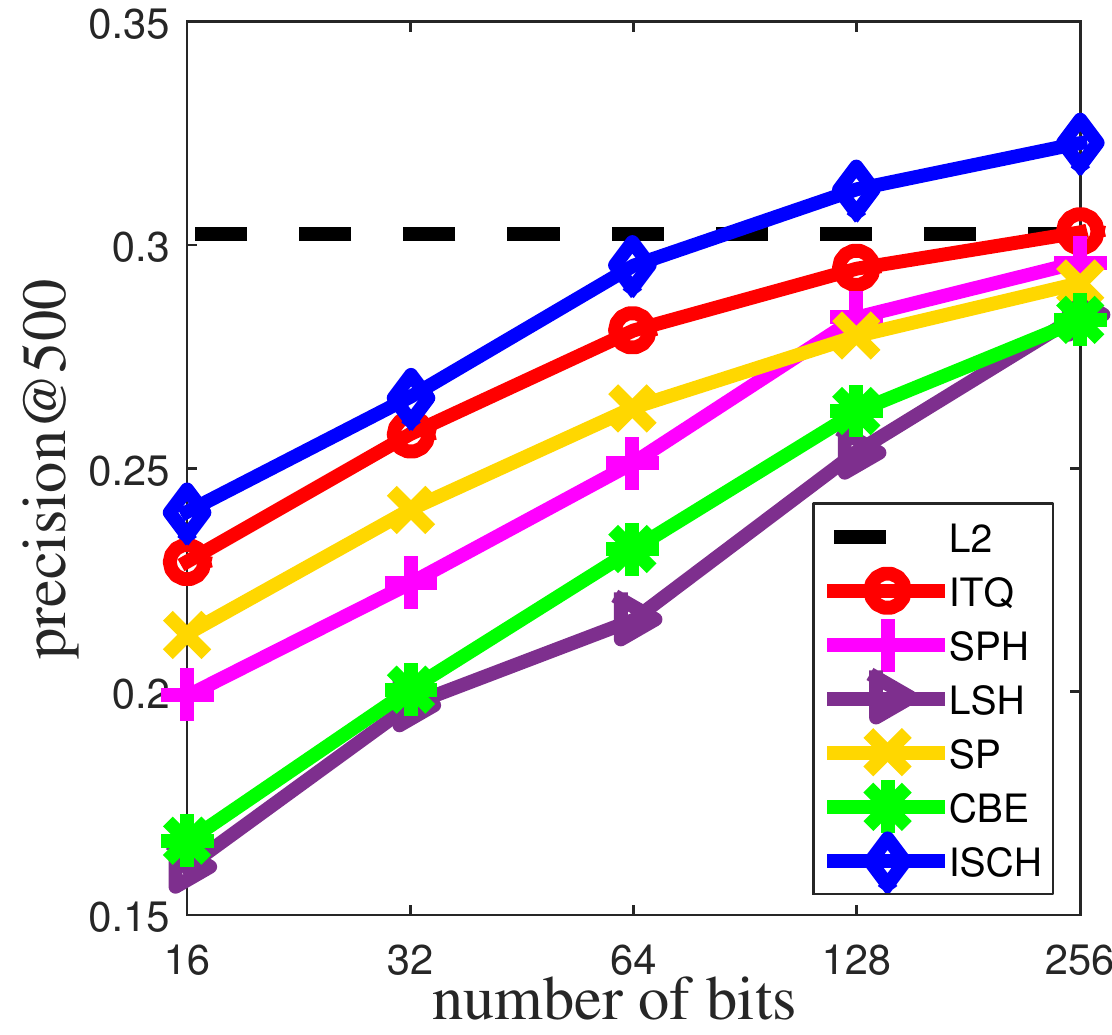} &
\includegraphics[width=0.32\textwidth]{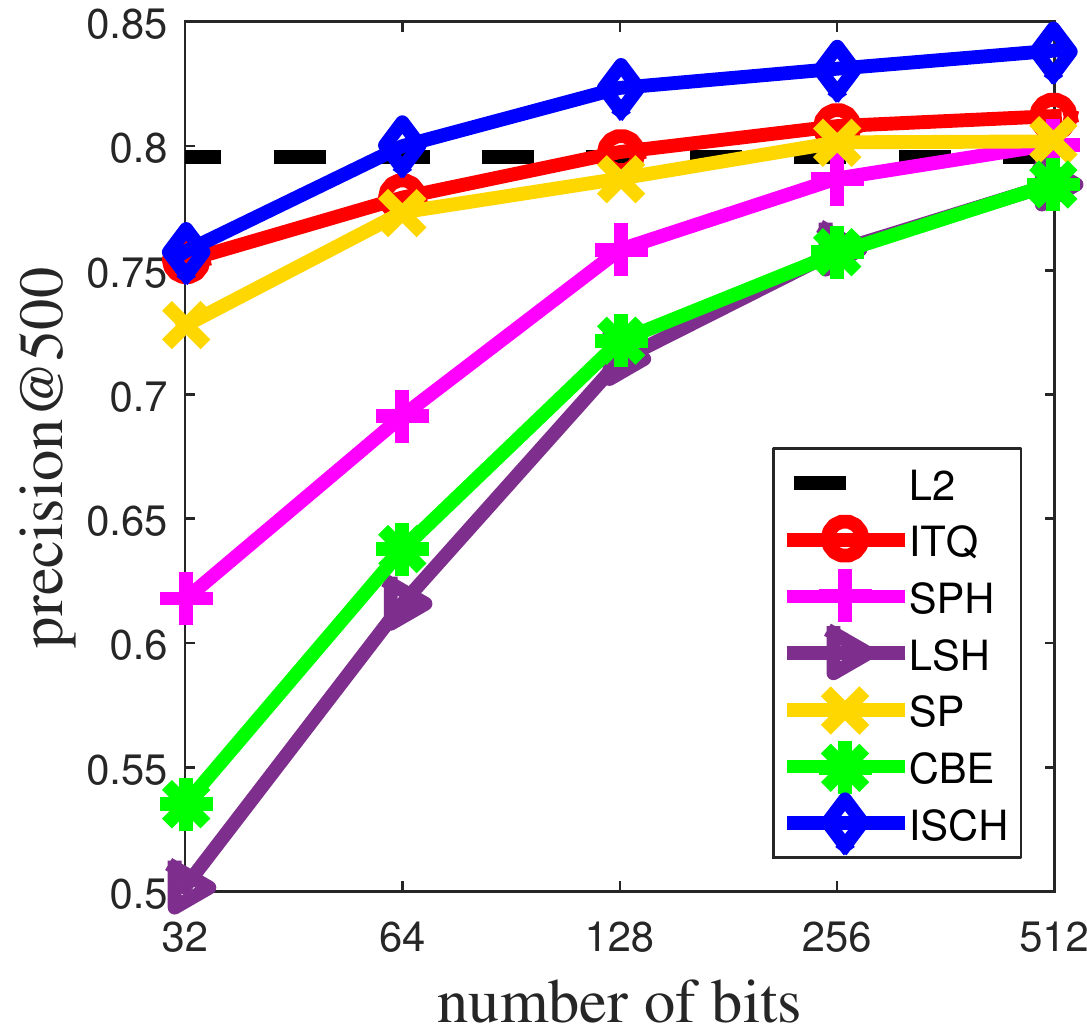} &
\includegraphics[width=0.32\textwidth]{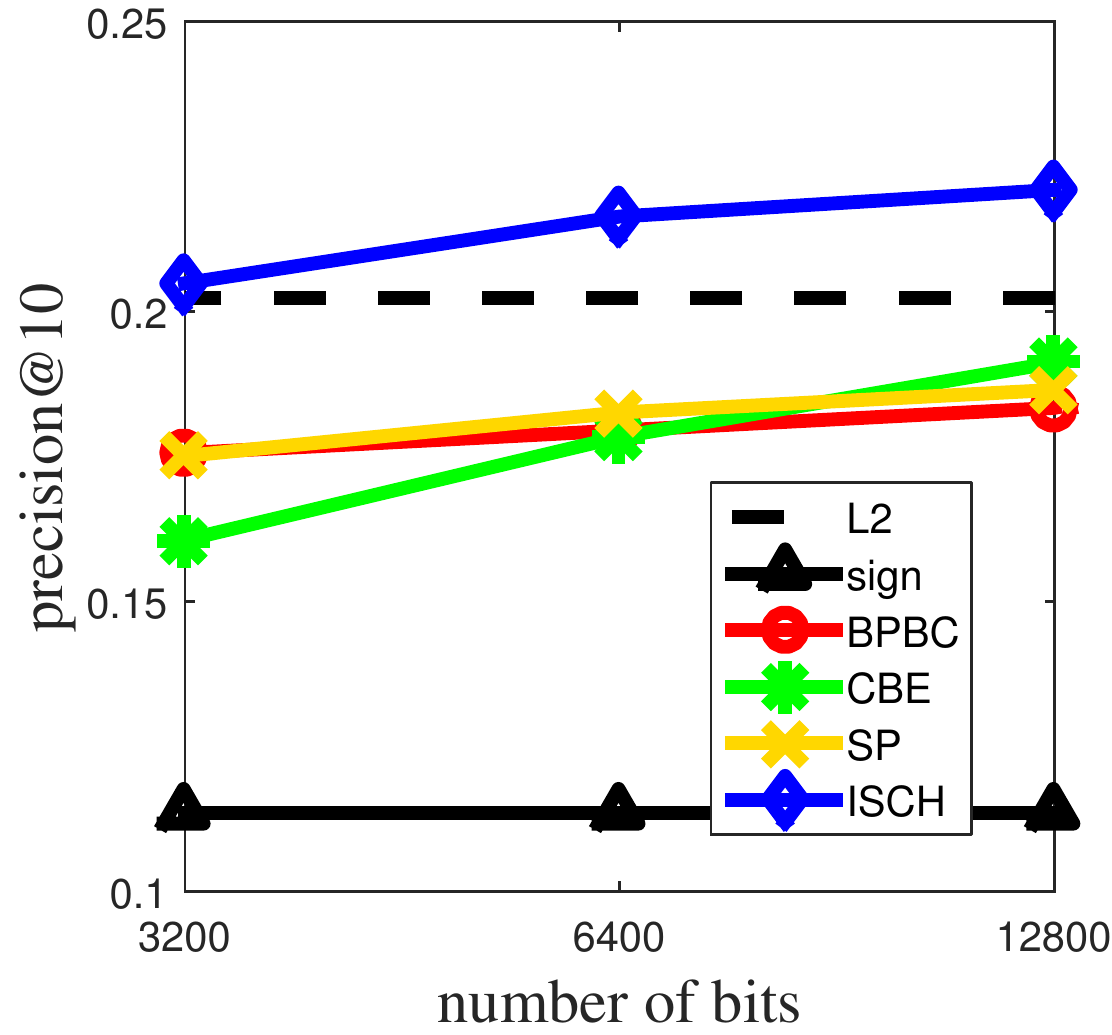} \\
(a) CIFAR-10 & (b) MNIST & (c) ILSVRC-VLAD \\ \\
\includegraphics[width=0.32\textwidth]{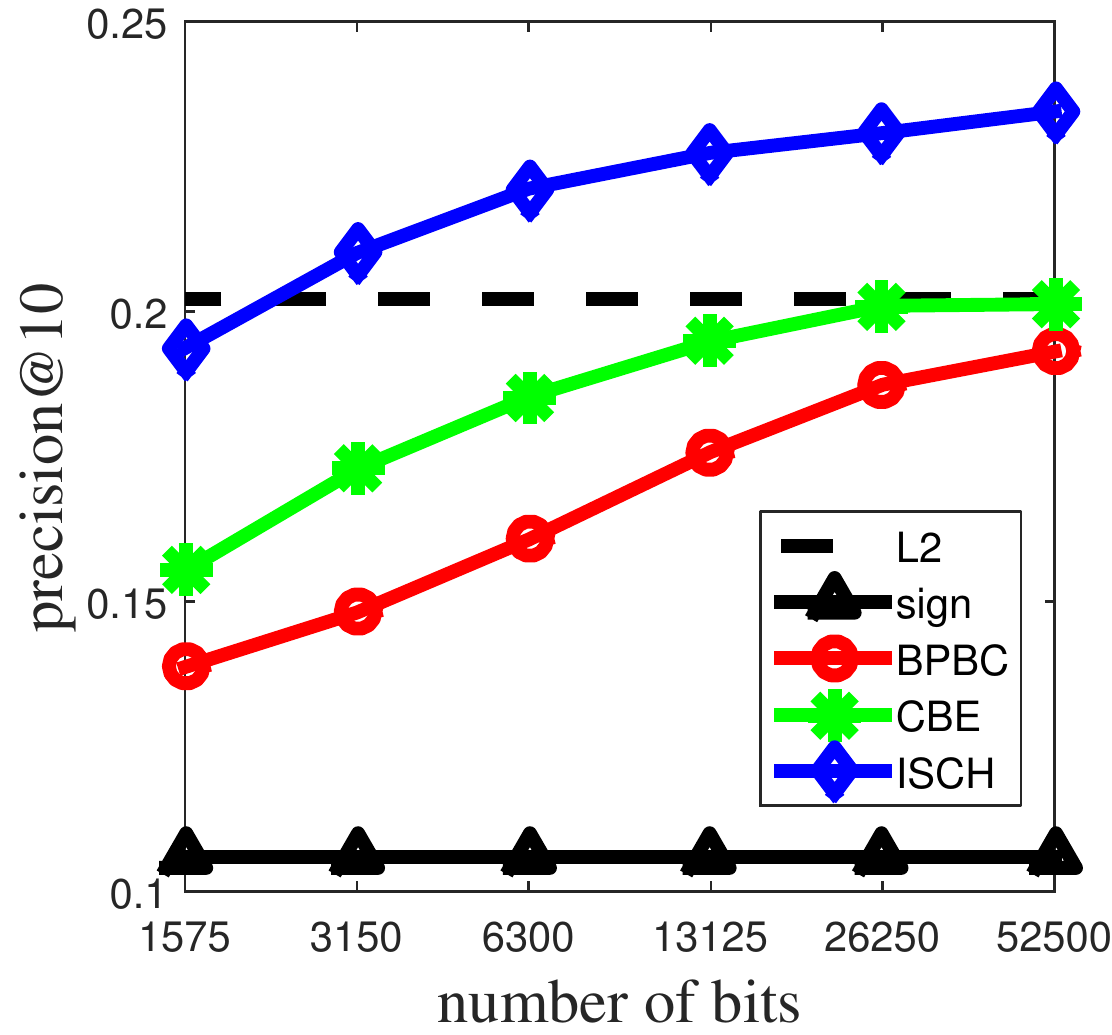} &
\includegraphics[width=0.32\textwidth]{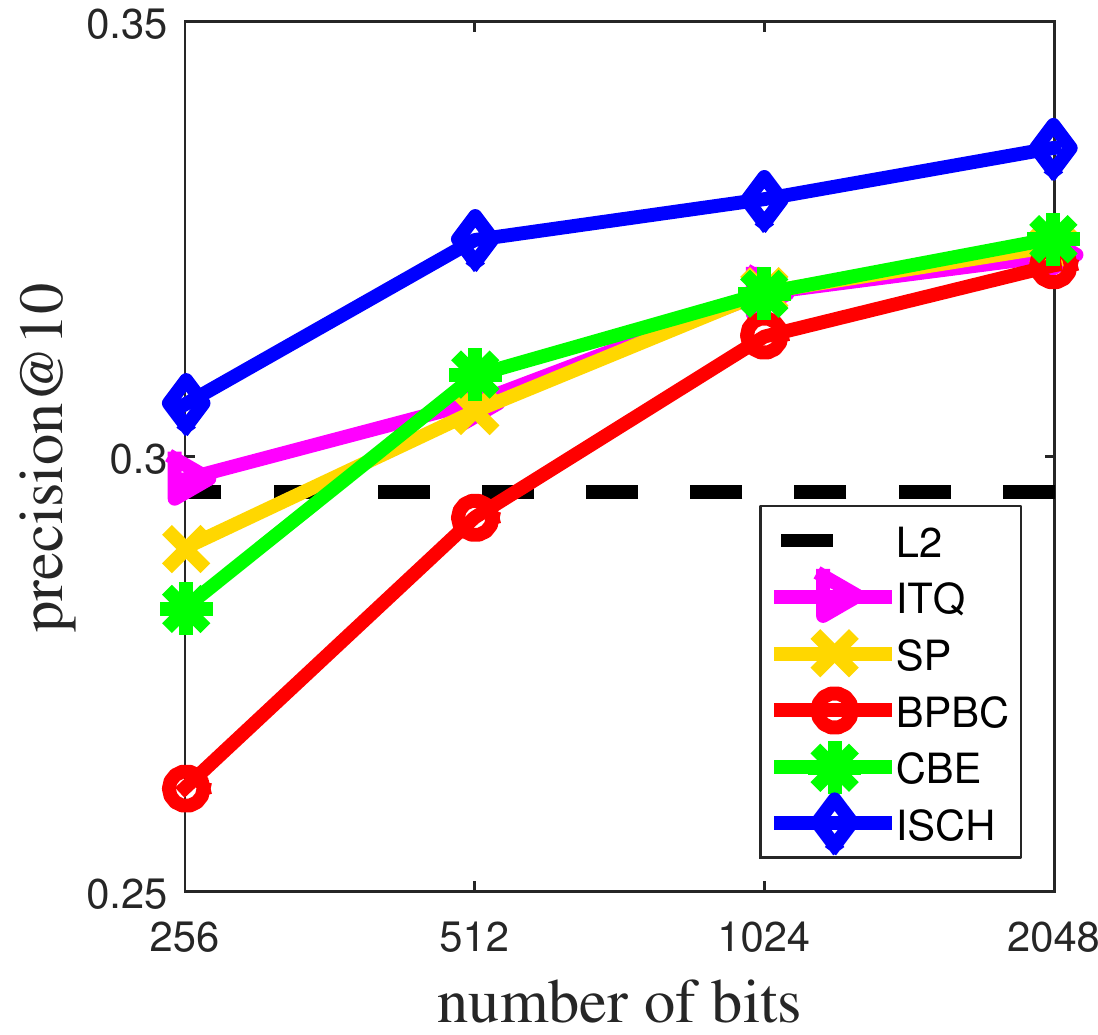}&
\includegraphics[width=0.32\textwidth]{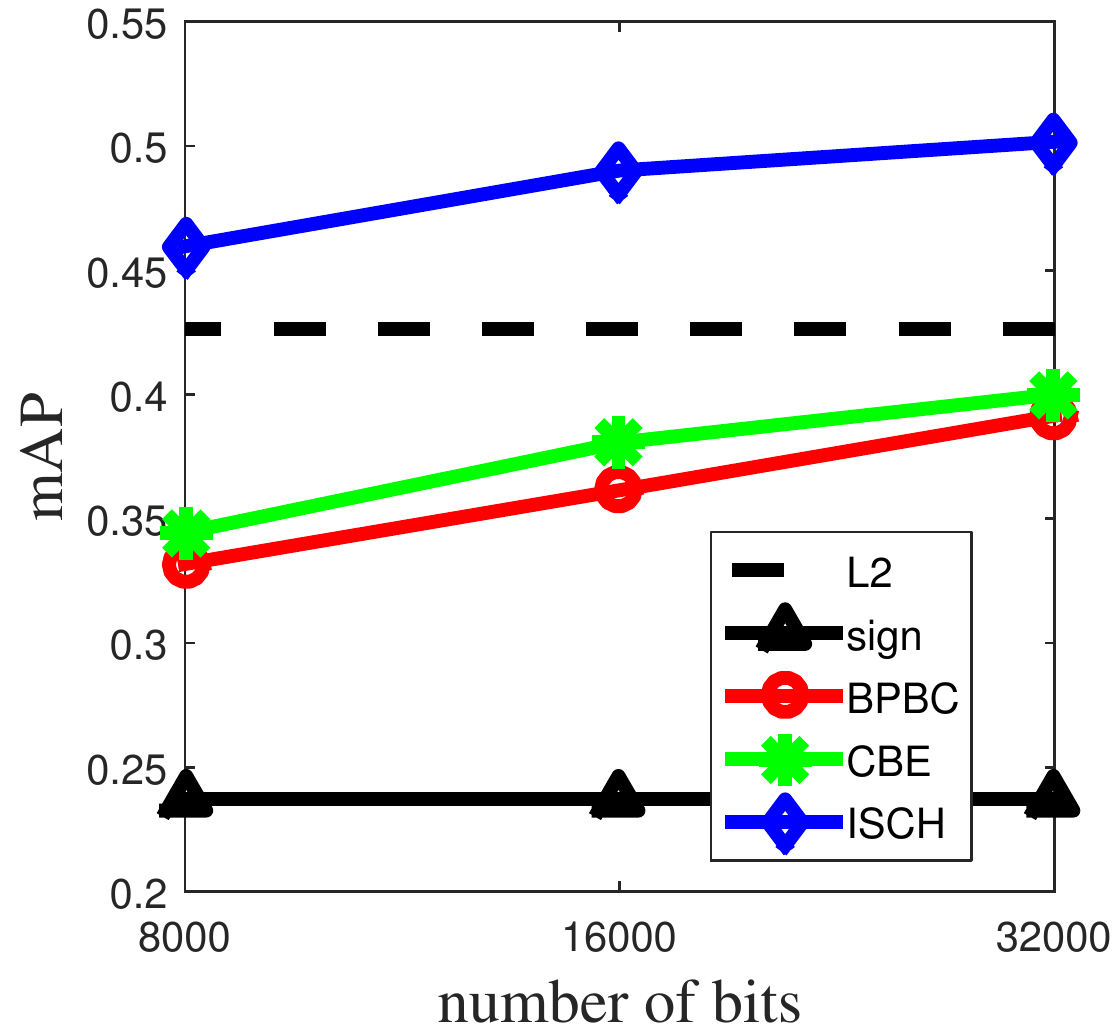} \\
(d) ILSVRC-LLC & (e) ILSVRC-CNN-fc7 & (f) Holiday+Flickr1M \\ \\
\includegraphics[width=0.32\linewidth]{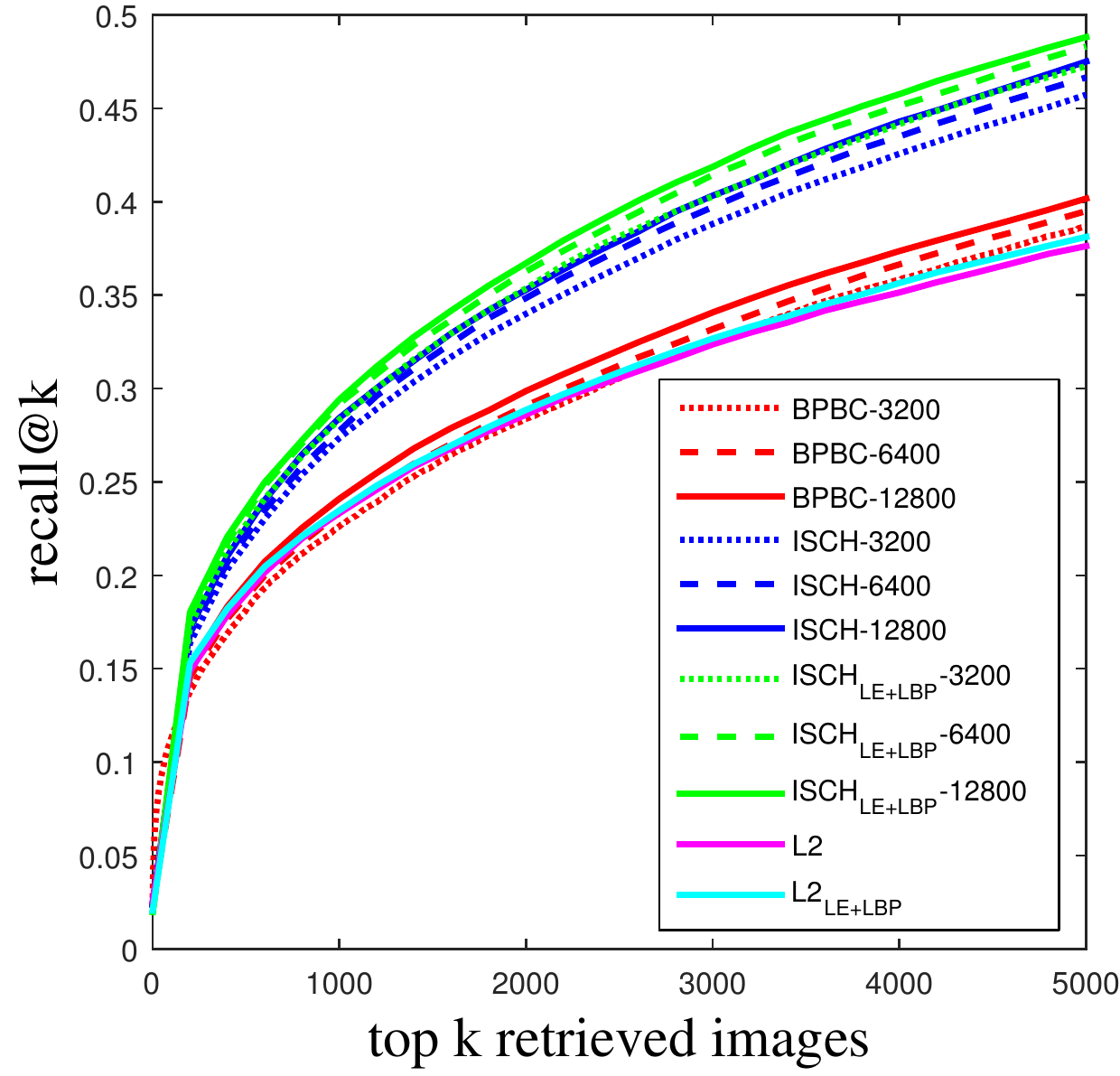} &
\includegraphics[width=0.32\textwidth]{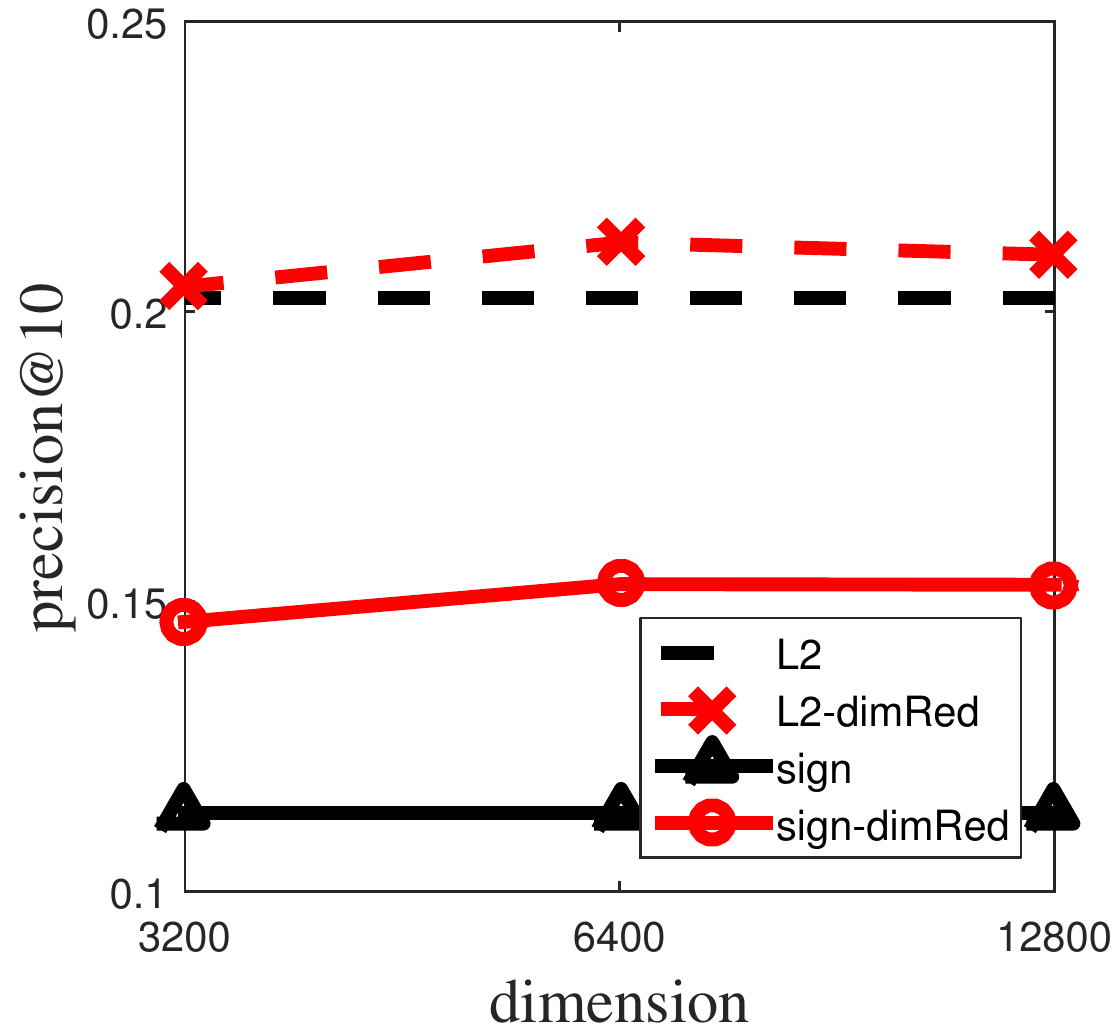} &
\includegraphics[width=0.32\textwidth]{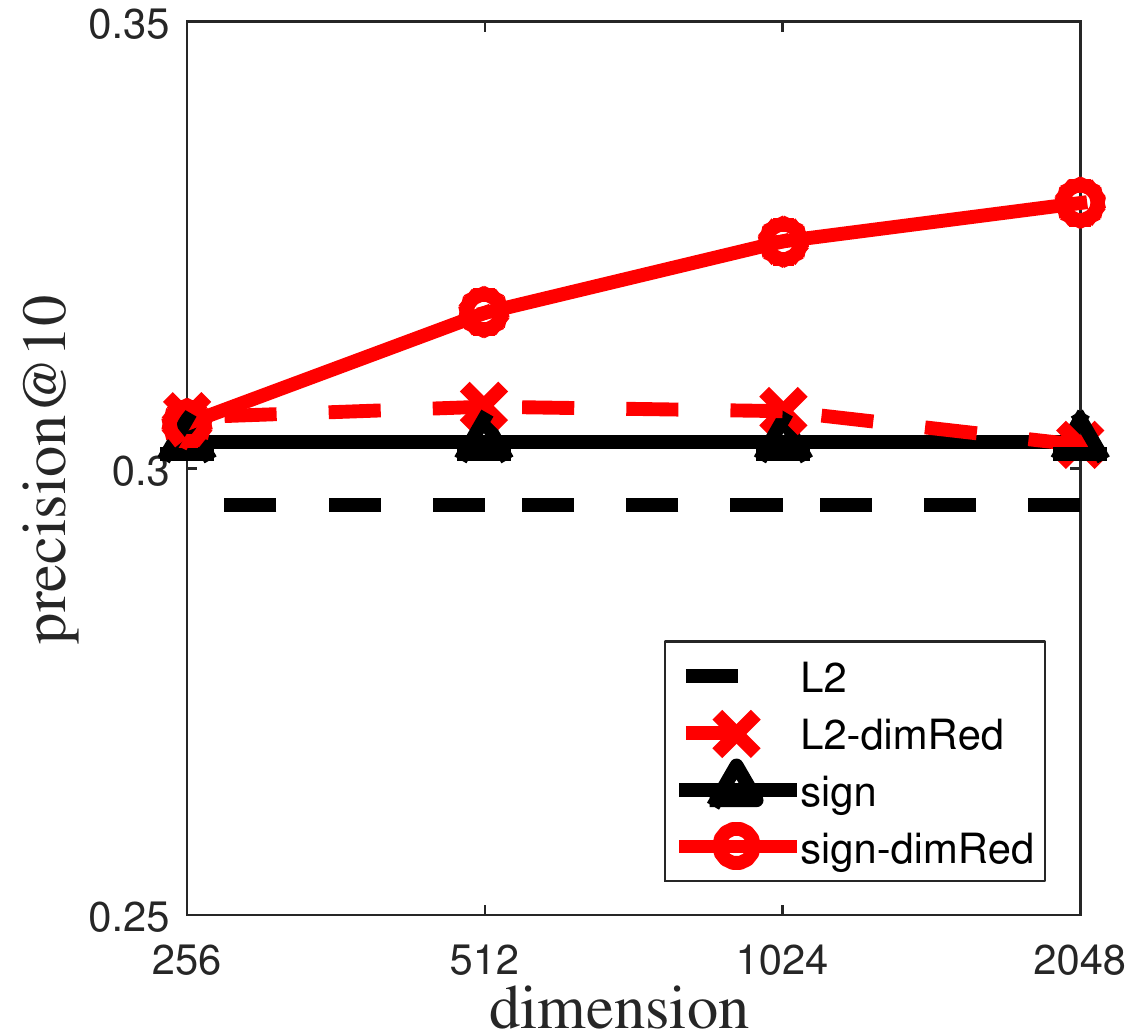} \\
(g) WDRef+LFW & (h) ILSVRC-VLAD-DR & (i) ILSVRC-CNN-DR \\
\end{tabular}
\caption{(a)-(e) Precision at k versus code length for CIFAR-10, MNIST, ILSVRC using $25600$-D VLAD, ILSVRC using $105000$-D LLC and ILSVRC using $4096$-D CNN-fc7, respectively. (f) Mean average precision versus code length for Holiday+Flicke1M using $64000$-D VLAD. (g) Retrieval performances on WDRef+LFW, measured by average recall of top k returns. The number following a specific method is the code size. "LE+LBP" indicates that face images are represented by the LE+LBP feature, otherwise, LE feature only. In (h) \& (i), the naive binary coding is decided by signs of the reduced features. The plots show precision at 10 for $2$-norm and Hamming distance retrievals versus reduced dimensions for $25600$-D VLAD and $4096$-D CNN-fc7. Notice that the results of performing $2$-norm and Hamming distance retrievals in the original input space are plotted in black. }
\label{fig:result}
\end{figure*}

\subsection{Retrieval results}
\label{ssec:results}

We begin with the experiments on  CIFAR-10 and MNIST. Evaluations are done by returning the top 500 ranked images of each query and reporting the average precision. Figures~\ref{fig:result}(a) and \ref{fig:result}(b) show the averaged precision of 500 nearest neighbors returned by each method. While all the compared binary coding techniques are designed to respect the $2$-norm  image relationships in the original feature space, our method is to preserve those implied by the inner products of sparse codes. The advantage is manifest in that a 128-bit binary code by ISCH already outperforms all other implementations using more bits. We also see that most of the compared binary codes would gradually approach the $2$-norm performance when increasing their code size. Notice that we have set the sparsity parameter in SP to $0.9$ and the value of $\lambda$  in CBE to 1 in all our experiments.

For ILSVRC2010, we use three different image descriptors, $25600$-D VLAD, $105000$-D LLC and $4096$-D CNN-fc7 . Before we report the evaluation results of the various techniques, we first show that the retrieval accuracy based on these sophisticated descriptors can still be boosted through performing dimensionality reduction. (We consider only VLAD and CNN-fc7 in that the LLC dimension is too large to run PCA.) We respectively conduct 2-norm and Hamming distance search based on a naive sign-thresholding binary coding, and compare their performances before and after performing PCA. The retrieval outcomes support that improvements can be achieved through appropriate dimensionality reduction as shown in Figures~\ref{fig:result}(h) and \ref{fig:result}(i). We then proceed to investigate the effectiveness of different binary coding schemes. When images in ILSVRC2010 are encoded by a $25600$-D VLAD, we apply the hierarchical k-means clustering described in the previous section to obtain a dictionary of size 32,385. For each method, we generate binary codes of size in half (12,800), quarter (6,400), and eighth (3,200) of the feature dimension. To achieve the best retrieval performance by BPBC, we have tried several combinations of the reduced dimensions of a bilinear projection, except for the code size 12,800, where we adopt the specific reduced dimensions reported in~\cite{GongKRL13}. Figure~\ref{fig:result}(c) shows the precision results at 10 retrieved images using binary codes of different lengths. The experiment for ILSVRC2010 with $105000$-D LLC is performed analogously. We construct a dictionary of size 12,9281. Despite that the dimensions of $DD^T$ now grow to $105000 \times 105000$, we can still use the approximation techniques to compute the binary codes. In Figures~\ref{fig:result}(c) and ~\ref{fig:result}(d), $3200$-bit ISCH already yields better performances than all other techniques with substantially large code sizes. In testing ISCH with the $4096$-D CNN-fc7 feature, the dictionary size is set to 8,000. Since the Caffe CNN-fc7 yields supervised features, the retrieval performances of all methods, plotted in in Figure~\ref{fig:result}(e), are significantly improved.

The image descriptor used in the INRIA Holiday+Flickr1M dataset is $64000$-D VLAD. For each query, there are about two relevant images in the database to be retrieved. We identify the retrieved images according to Hamming distances and compute the average precision for each query. For ISCH, we construct a dictionary of size 77,001 to generate binary codes of 8,000, 16,000 and 32,000 bits. Figure~\ref{fig:result}(f) includes the mAP curve for different code sizes of each method. The results show that ISCH uniformly outperforms BPBC and CBE by appreciable margins.

In many challenging computer vision problems, adopting a combined feature vector by fusing different image descriptors is often useful. In the experiment with WDRef+LFW, we demonstrate that ISCH is flexible in this respect. In particular, ISCH only requires that the target code length can be factorized by $m = Q \times \ell$. However, such a generalization is not applicable to BPBC in that it relies on the two-dimensional structure of a high-dimensional descriptor. Since not all image descriptors have an implicit two-dimensional structure (or even they do, the structures are generally not the same), the feature fusion could not be a workable option for BPBC. To describe each face image in WDRef+LFW, we use LE and LBP descriptors, where the latter does not display a two-dimensional structure. We consider two feature representations based on LE and LE+LBP, and construct two dictionaries of size 34,835 and 34,571, respectively. We generate binary codes of size 3,200, 6,400, and 12,800 for both methods and plot the results measured by recall value in Figure~\ref{fig:result}(g). We see that ISCH considerably outperforms BPBC when working with the LE feature only. The average recall value of ISCH at top 5,000 returned images is over 7\% more than that of BPBC. Furthermore, in Figure~\ref{fig:result}(g), the retrieval performances by ISCH can noticeably benefit from considering the combined feature LE+LBP, while BPBC is not applicable.

%
\section{Discussions}
\label{sec:con}
%

Previous uses of sparse codes to better define image structure \cite{timofte2011sparse} or index a hash table for similarity search \cite{CherianSMP14} require explicitly solving them. However computing sparse codes for large-scale high-dimensional data is not only extremely time-consuming (or even unfeasible) but also parameter sensitive. The technique developed in \cite{GkioulekasZ11} provides a convenient way of skipping such computation in finding a dimensionality-reduction mapping that preserves the inner products of sparse codes. Since it is achieved by minimizing the {\em expected} squared difference of inner products under the assumption of a sparse linear model, the resulting mapping is reasonably robust if the underlying rotation matrix can be properly constructed. In our formulation, we link the criterion of an optimal rotation matrix to yielding good binary codes that minimize the quantization error. The key to our approach for tackling large-scale computation is the use of {\em problem decomposition}. By coupling approximate spectral decomposition and the assumption of rotation matrix being sparse block-diagonal, we reduce the daunting eigenproblem and the quantization-error minimization into independent and manageable subproblems. All these efforts result in a new and promising technique, coined as ISCH, to generate effective binary codes for large-scale image data.

We conclude by briefly discussing the complexity of ISCH. Regarding the time complexity, excluding the step of learning a dictionary, the time complexity for training is $O(d\ell^2+n\ell^2)$ and for computing the binary code of a testing image is $O(dm)$. For the storage usage of ISCH, we take BPBC as a comparative model. From the experimental results, we observe that our method needs only at most one fourth of bits in a binary code as BPBC to achieve comparable retrieval performances. Using ILSVRC2010 with $25600$-D VLAD  as an example, $3200$-bit ISCH would need extra storage of $25600 \times 3200 \times 4$ (byte) $\sim  328$MB for the dimensionality-reduction mapping.  $12800$-bit BPBC requires  $12800 \times 1.2M \div 8 \sim 1.9$GB for storing the binary codes of the whole dataset, while our method only needs $1.9\times 0.25=0.475$GB for the codes.  Nevertheless, as the feature dimension $d$ increases, the advantage of storage saving by ISCH would diminish since the storage for the mapping grows up in $O(d^2)$. This suggests that how to approximate the learned $L^*$ in (\ref{eqn:L}) to save the storage is essentially a practical research topic for future direction.

{\small
\bibliographystyle{ieee}
\bibliography{ISCH}
}

\end{document}